# High-Resolution Vision Transformers for Pixel-Level Identification of Structural Components and Damage

Kareem Eltouny[1], Seyedomid Sajedi[1], and Xiao Liang[1]

**ABSTRACT**

Visual inspection is predominantly used to evaluate the state of civil structures, but recent developments in unmanned aerial vehicles (UAVs) and artificial intelligence have increased the speed, safety, and reliability of the inspection process. In this study, we develop a semantic segmentation network based on vision transformers and Laplacian pyramids scaling networks for efficiently parsing high-resolution visual inspection images. The massive amounts of collected high-resolution images during inspections can slow down the investigation efforts. And while there have been extensive studies dedicated to the use of deep learning models for damage segmentation, processing high-resolution visual data can pose major computational difficulties. Traditionally, images are either uniformly downsampled or partitioned to cope with computational demands. However, the input is at risk of losing local fine details, such as thin cracks, or global contextual information. Inspired by super-resolution architectures, our vision transformer model learns to resize high-resolution images and masks to retain both the valuable local features and the global semantics without sacrificing computational efficiency. The proposed framework has been evaluated through comprehensive experiments on a dataset of bridge inspection report images using multiple metrics for pixel-wise materials detection.

**INTRODUCTION**

The recent breakthroughs in computer hardware and sensing technology have provided tools that facilitate the collection of high-quality visual media during structural inspections. The large amount of obtained footage, however, is often examined manually by the inspectors for structural defects which could be challenging to perform if a large urban area was affected by a major hazard. To address this problem, many studies focused on developing an automated visual inspection algorithm usually involving morphological operations, edge detection algorithms, filtering, binarization, and other image processing techniques [1, 2]. However, in recent years, significant progress has been made in the field with the advent of artificial intelligence-based

[1] University at Buffalo, the State University of New York, Buffalo NY 14260, USA

computer vision algorithms. These techniques can leverage large datasets to learn meaningful features enabling more accurate and efficient defect detections when compared to conventional visual inspection methods.

Recent advances in deep learning have, without a doubt, impacted structural health monitoring (SHM) [3]. Many methodologies have been proposed over the past few years that utilize deep and complex neural network architectures for automating structural condition assessment procedures including convolutional neural networks (CNNs) [4], generative adversarial networks [5], autoencoder neural networks [6], recurrent neural networks [7, 8], and hybrid deep learning models,[9]. It is hence no surprise that deep learning has also revolutionized automated visual inspection, enabling more accurate and efficient defects detections in many applications. There are two main directions in developing vision-based SHM: component detection and damage detection. Detection of structural and non-structural components and materials often has the purpose of accelerating the inspection process such as providing guidance to the inspectors or the deployed UAVs. Multiple methods were proposed in the past using various techniques including object detection [10, 11], and semantic segmentation [12, 13]. The bulk of the vision-based SHM research, however, was dedicated to structural damage detection including concrete cracks [13, 14], pavement cracks [15], and building façades [16]. Multi-task methods combining multiple visual inspection tasks were also proposed [17].

It is common to adopt established deep learning algorithms for developing vision-based SHM using transfer learning. But it is worth noting that structural inspections, a procedure of high-stakes nature, often requires high precision and accuracy achieved by using input data of relatively high resolution. In addition, many visual inspection tasks require real-time inference which can, for example, be used for UAV navigation. On the other hand, state-of-the-art computer vision models, such as vision transformers (ViT), are becoming increasingly expensive in terms of computational demands which makes the use of high-resolution data even more challenging. Two models were proposed recently to address these challenges by using different strategies for handling high-resolution images efficiently [18]. However, a single approach was not successful in addressing the needs of different visual inspection tasks.

In this study, we develop a unified framework for high-resolution visual inspection that can strike a balance between prediction quality and computational efficiency. Uniform downsizing of images, which is commonly performed when resources are limited, can distort the original image and cause a loss of fine details. We propose Swin transformer segmentation with trainable resizers (SwinTR), a transformer-based segmentation model paired with cascaded sub-pixel convolution scaling networks. SwinTR brings the ViT technology into the field of vision-based SHM with insignificant increase in computational costs. We evaluate the framework on the Structural Materials Segmentation dataset, but it can be applied to other visual inspection tasks with minor adjustments.

**ARCHITECTURE DESIGN**

Many off-the-shelf semantic segmentation models are optimized for low to medium-resolution images. Using a state-of-the-art segmentation model on high-resolution images could be met with GPU-memory constraints making it highly inefficient. To handle high-resolution images, a common practice is to uniformly downsample the images before feeding them into the segmentation model to cater to the available computational resources. However, this process often impacts the segmentation performance especially near object boundaries. Another way to handle this, especially for crack damage detection, is to dissect the image into smaller parts in a process known as patch cropping at the expense of losing global semantic information. Instead, and inspired by super-resolution research, we propose two neural networks that can allow using a low-resolution segmentation model on high-resolution datasets with a relatively low increase in computational demands.

Our SwinTR network includes two main encoder-decoder stages: the trainable resizers in the outer stage, and an internal, low-resolution Swin Transformer-based U-Net++ segmentation model (Figure 2). We find this design to have high computational and memory efficiency compared to using high-resolution segmentation models. For example, we estimated an approximate 8.4 GB of GPU memory demand to pass a single 1920×1080 image through a U-Net model compared to 1.6 GB needed using our approach.

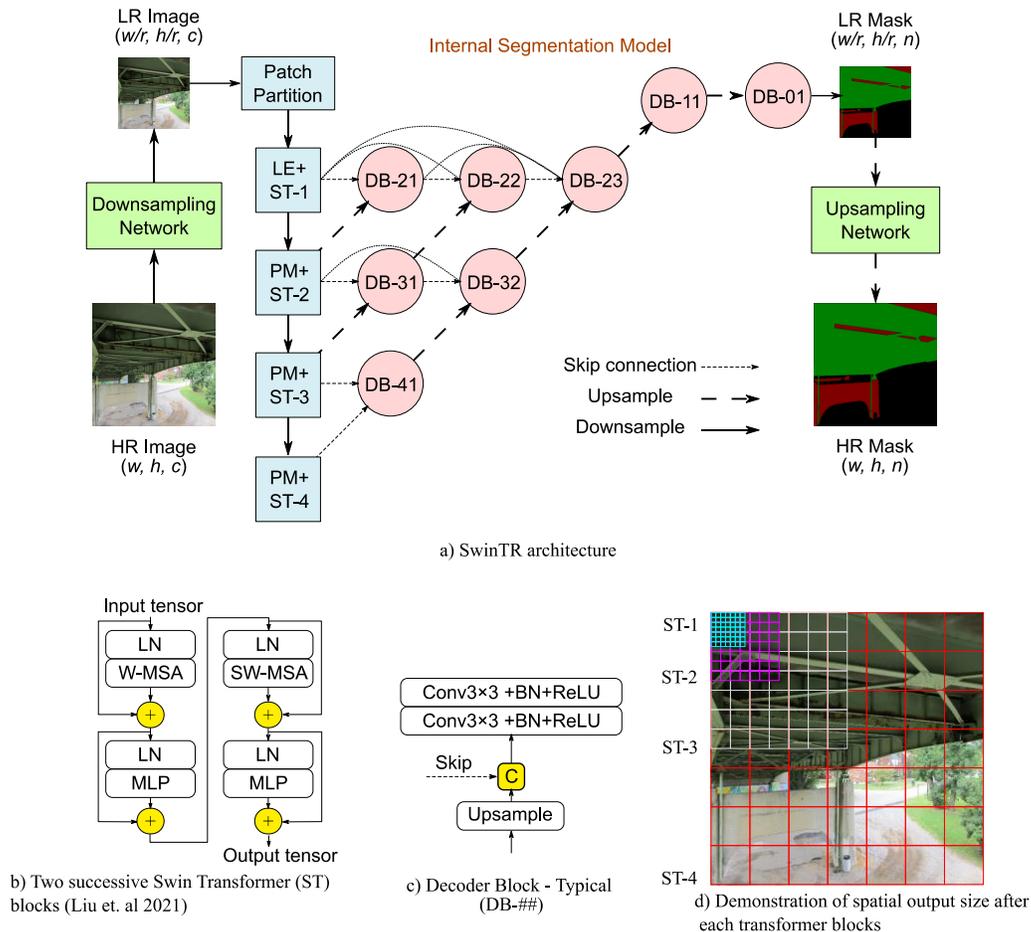

Figure 2. SwinTR architecture ($c$: channels, $n$: classes, ST-$i$: Swin Transformer block, D-$i,j$: decoder block, Conv: 2D convolution, BN: batch normalization, ReLU: rectified linear unit, LN: Layer normalization, MLP: multi-layer perceptron, SW/W-MSA: regular and shifted windowed multi-head self-attention, LE: linear embedding, PM: patch merging [19]).

The concept behind the use of the trainable resizers (Figure 3) is that, unlike uniform sampling where all pixels have equal importance, some pixels matter more than others. The downsampler and upsampler networks, therefore, learn to sample valuable pixels more frequently in the resizing process. The two symmetric networks are inspired by deep Laplacian pyramid networks which were used in the past for super-resolution and image generation methods [20, 21]. The upsampling network, called Laplacian Subpixel Convolutional Network (LapSCN), contains a feature extraction branch and a mask reconstruction branch. The downsampling network, called Laplacian Desubpixel Convolutional Network (LapDCN) replaces the reconstruction branch with an image deconstruction branch. The two networks can progressively rescale images or masks based on a cascade of subpixel convolutional modules. At each level, the data is fed into the feature extraction branch which is responsible for learning the higher resolution residuals. The residuals can then be used to fine-tune the resized data in the deconstruction/reconstruction branch through addition. The result from the deconstruction or reconstruction branch can be either used as the resized output or fed into the feature branch for an additional stage of scaling.

The feature extraction branch in LapSCN contains subpixel convolutional blocks [22] where each block provides residuals for masks upsampled by a factor of two. The final pixel shuffle layer reorganizes the low-resolution feature maps to form two-times upscaled feature maps. The desubpixel convolutional blocks, which are used for feature extraction in LapDCN, begins instead with a pixel unshuffle layer, a reverse operation to pixel shuffle. These layers can drastically reduce the information loss during downsampling as pixels are rearranged into the channels' axis.

The internal segmentation model uses a state-of-the-art Swin Transformer backbone [19] with a U-Net++ decoder [23]. Transformers, which were originally used in natural language processing [24], have been dominating many computer vision benchmarks in recent years including ImageNet and ADE20k [25-27]. It is, however, not straightforward to use transformer networks as encoders in a U-Net-like segmentation architecture. Therefore, Swin Transformer relies on patch merging operations to provide hierarchical feature maps at each stage of the network. We have fitted the decoder to a Swin Transformer Base (Swin-B) model that is pretrained on ImageNet with an input/output resolution of 224×224.

**CASE STUDY: MATERIAL SEGMENTATION DATASET**

The material segmentation dataset is a publicly available dataset containing 3,817 images extracted from the Virginia Department of Transportation bridge inspection reports [28] with pixel-level annotation of three structural inspection materials: concrete, steel, and metal decking. The images resolution varies from a low of 256×237 to a high of 5184×3456, making it a suitable test bench for our framework. The dataset is split into 3,436 samples for training and 381 samples for testing by its original authors. We further set aside 352 samples from the training dataset to be used for validation during network training (approximately 10% of the training set).

The models were built and trained using the PyTorch library [29] and a workstation equipped with Intel® Core i9-13900k CPU and an Nvidia RTX 4090 GPU. Multiple data augmentation techniques were carried out during training including a variety of random color manipulation and image transforms to help reduce training overfitting. We used the focal loss [30] as the cost function and Adam as the optimizer [31] in addition to learning rate schedulers having maximum and minimum learning rates of 1E-4 and 1E-6, respectively. All models were trained for 50 epochs and the selected checkpoint corresponds to weights resulting in the maximum intersection-over-union (IoU) of the validation set.

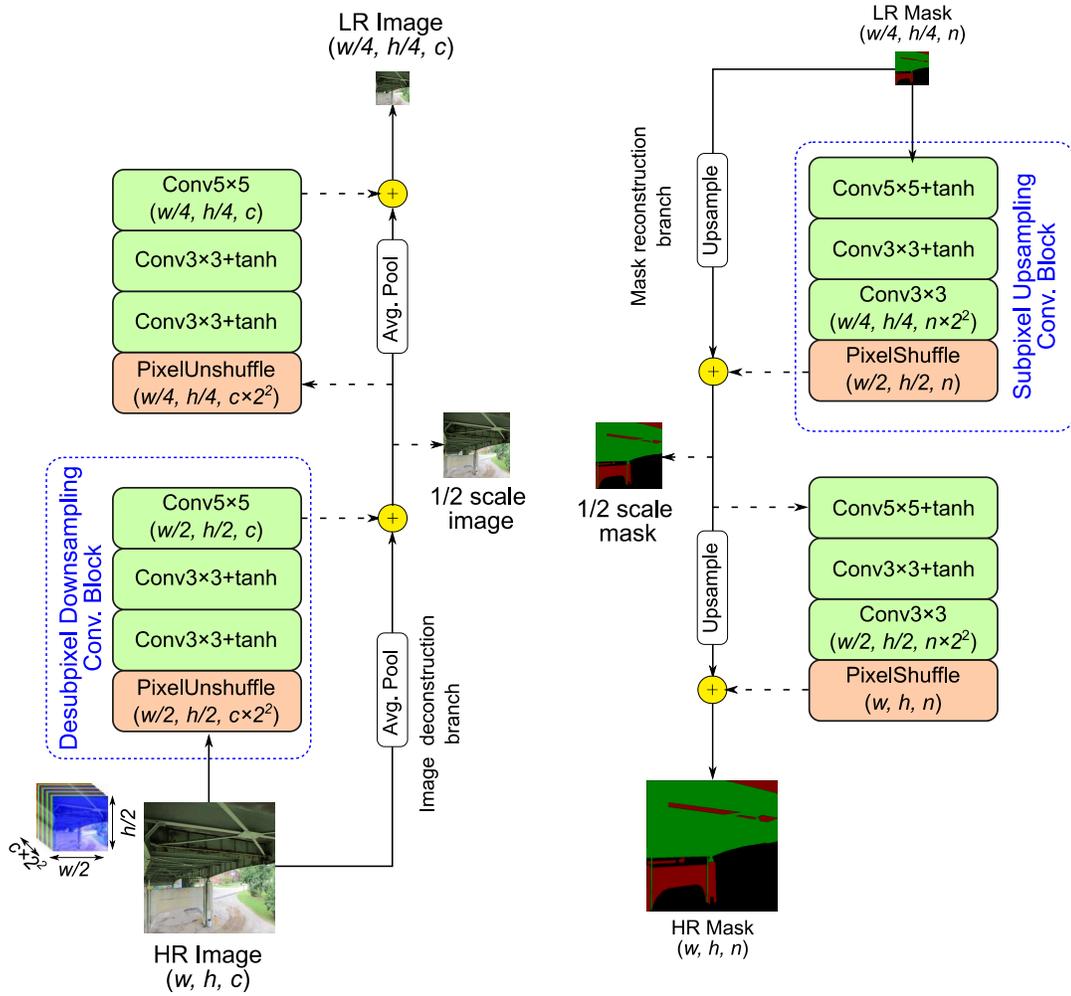

a) LapDCN architecture (downsampling network)          b) LapSCN architecture (upsampling network)

Figure 3. Downsampling and upsampling network architectures (*c*: channels, *n*: classes, Conv: 2D convolution, BN: batch normalization, ReLU: rectified linear unit).

## RESULTS

Five segmentation models with different input sizes are compared using the material segmentation dataset based on four metrics: precision, recall, F1-score, and IoU. All models share the same internal segmentation model based on Swin Transformer. The internal SwinTR model is trained for images and masks of sizes 224×224. The uniform SwinTR attaches interpolation-based upsampler and downsampler to the internal model downstream and upstream which allows it to handle images and masks four times the size (896×896). SwinTR 2× and SwinTR 4× are two SwinTR variants that elevate the sizes of the segmentation model by two and four times, respectively. The average results of all these five models are shown in Tables I and II. Overall, using the trainable resizers has allowed for a slight increase in performance in all metrics with SwinTR 2× achieving the best results. Using non-trainable resizers, however, slightly deteriorated the model performance when compared to the low-resolution segmentation model. It is worth noting that the differences in performance are subtle and, for this dataset, the use of interpolation-based resizers should not drastically impact the prediction results.

Figures 4 and 5 show the output of LapDCN and LapSCN alongside the outputs of non-trainable resizers. In Figure 4, it is observed that uniformly downsampling the image can distort the edges which can be valuable for defining the boundaries of the segmentation object. These boundaries and other important details are better retained in the LapDCN output in a low-resolution form. A similar conclusion can be made by upscaling the masks (Figure 5). LapSCN can learn to provide a super-resolution version of the supplied masks by the internal segmentation model compared to the interpolation-based upsampler. In addition, the boundaries of the segmentation objects show clear artifacts compared to the smoothed boundaries of LapSCN masks.

TABLE I. TESTING PERFORMANCE METRICS (AVERAGE)

|  | Size | Precision (%) | Recall (%) | F1-score (%) | IoU (%) |
|---|---|---|---|---|---|
| Internal SwinTR | 224 | 92.65 | 92.09 | 92.35 | 86.05 |
| Uniform SwinTR 4× | 896 | 92.45 | 91.90 | 92.14 | 85.72 |
| SwinTR 2× | 448 | **92.94** | **92.02** | **92.44** | **86.25** |
| SwinTR 4× | 896 | 92.78 | 91.99 | 92.35 | 86.07 |

TABLE II. CLASS-WISE INTERSECTION-OVER-UNION RESULTS

|  | Background | Concrete | Steel | Metal decking |
|---|---|---|---|---|
| Internal SwinTR | 74.97 | 86.46 | 94.15 | 88.63 |
| Uniform SwinTR 4× | 74.28 | 86.56 | 94.06 | 87.97 |
| SwinTR 2× | 74.26 | **87.14** | **94.49** | **89.09** |
| SwinTR 4× | **74.57** | 86.64 | 94.20 | 88.87 |

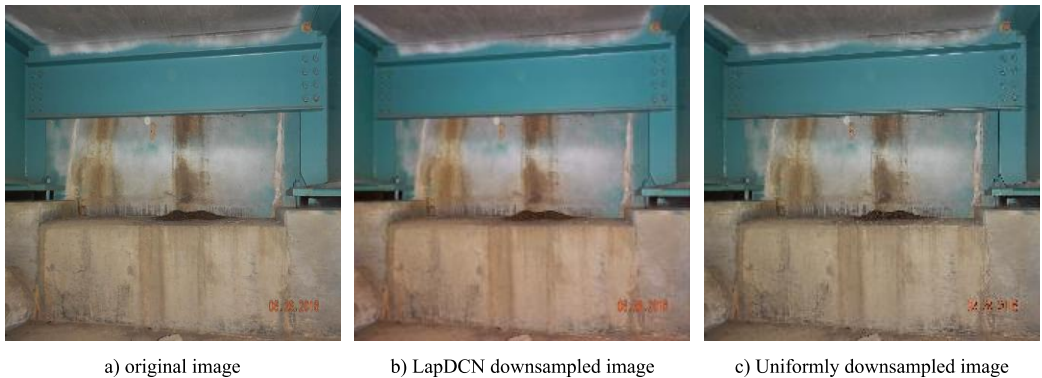

a) original image　　　　b) LapDCN downsampled image　　　　c) Uniformly downsampled image

Figure 4. Different image downsamplers results.

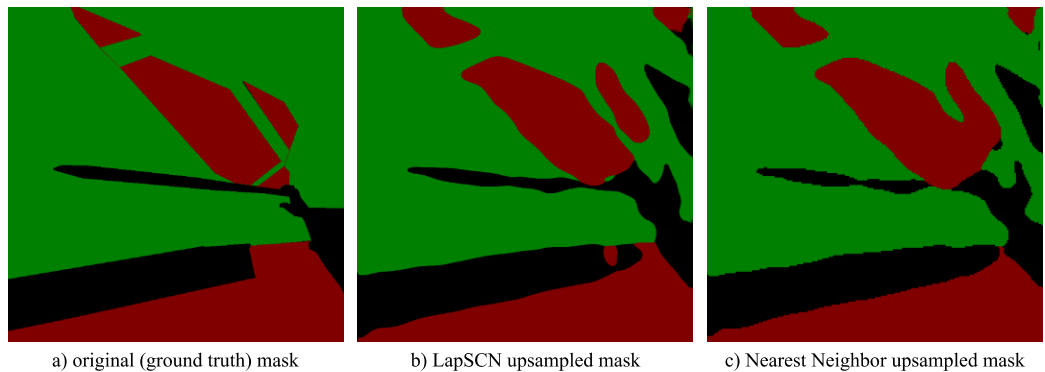

a) original (ground truth) mask     b) LapSCN upsampled mask     c) Nearest Neighbor upsampled mask

Figure 5. Different mask upsamplers results.

## CONCLUSIONS

As more progress is made in hardware and sensing technology, engineering inspectors gain access to an increasing amount of high-dimensional data that would benefit from the advances in artificial intelligence. Fast and efficient visual inspection is preferred for real-time applications, but the resource demands of state-of-the-art vision transformer models are exponentially increasing. We proposed a high-resolution visual inspection framework that encompasses a low-resolution transformer segmentation network and two trainable resizers inspired by efficient subpixel convolution and Laplacian pyramid networks. By testing our framework on the material segmentation dataset, we found that while there were gains in accuracy and IoU compared to the model with interpolation-based resizers, the increase is practically insignificant. We also found that the use of non-trainable downsamplers and upsamplers can impact the object boundaries negatively, which can be valuable for certain tasks. Future research will include testing the framework on tasks that are more sensitive to image resolution, such as crack segmentation.